# Design of one-year mortality forecast at hospital admission based: a machine learning approach

Authors: Vicent Blanes-Selva[1], Vicente Ruiz-García[2], Salvador Tortajada[2,3], José-Miguel Benedí[4], Bernardo Valdivieso[2], Juan M. García-Gómez[1]

1: BDSLab, Instituto de Tecnologías de la Información y Comunicaciones (ITACA), Universitat Politècnica de València, Camino de Vera s/n, 46022 Valencia, Spain;
2: Unidad Mixta de Tic aplicadas a la reingeniería de procesos socio-sanitarios ERPSS, Instituto de Investigación Sanitaria La Fe, Avenida Fernando Abril Martorell, 46026 Valencia, Spain;
3: Instituto de Física Corpuscular (IFIC), Universitat de València, Consejo Superior de Investigaciones Científicas, 46980 Paterna, Spain
4: Pattern Recognition and Human Language Technology (PRHLT) Research Center, Universitat Politècnica de València, Camino de Vera s/n, 46022 Valencia, Spain;

Abstract.

**Background:** Palliative care is referred to a set of programs for patients that suffer life-limiting illnesses. These programs aim to guarantee a minimum level of quality of life (QoL) for the last stage of life. They are currently based on clinical evaluation of risk of one-year mortality.

**Objectives:** The main objective of this work is to develop and validate machine-learning based models to predict the exitus of a patient within the next year using data gathered at hospital admission.

**Methods:** Five machine learning techniques were applied in our study to develop machine-learning predictive models: Support Vector Machines, K-neighbors Classifier, Gradient Boosting Classifier, Random Forest and Multilayer Perceptron. All models were trained and evaluated using the retrospective dataset. The evaluation was performed with five metrics computed by a resampling strategy: Accuracy, the area under the ROC curve, Specificity, Sensitivity, and the Balanced Error Rate.

**Results:** All models for forecasting one-year mortality achieved an AUC ROC from 0.858 to 0.911. Specifically, Gradient Boosting Classifier was the best model, producing an AUC ROC of 0.911 (CI 95%, 0.911 to 0.912), a sensitivity of 0.858 (CI 95%, 0.856 to 0.86) and a specificity of 0.807 (CI 95%, 0.806 to 0808) and a BER of 0.168 (CI 95%, 0.167 to 0.169).

**Conclusions:** The analysis of common information at hospital admission combined with machine learning techniques produced models with competitive discriminative power. Our models reach the best results reported in state of the art. These results demonstrate that they can be used as an accurate data-driven palliative care criteria inclusion.

## 1. Background and Significance

An increasing number of people have multiple morbidities and conditions in the final moments of their lives, current medicine tries to maintain a quality of life of these people, including their needs in the final moments. In this situation, palliative care tries to facilitate the life of people in these conditions from a patient perspective.

Palliative care is a multidisciplinary care that aims to grant comfort to the patient, avoid painful and/or aggressive treatments, alleviate pain, other symptoms, psychological and spiritual distress [11]. In addition, there are some studies which prove that patients receiving early palliative care present a better quality of life, mood, satisfaction with the treatment [8, 10, 12] and even a longer survival when compared to patients whose palliative care was delayed [9].

A criterion for the palliative care inclusion is desirable as early as possible. An adverse event such as a hospital admission could be considered a convenient episode to check this criterion. Nowadays, the main indicator to include a patient in palliative care is the clinical criterion of a potential exitus within the next 12 months. An example of that is the surprise question described in [21].

Mortality forecast has been previously studied by other groups. Buurman et al. in 2008 [1] proposed a method for predicting 90-days mortality risk using few clinical features: Barthel test index, Charlson score, Malignancy and Urea nitrogen (mmol/Liter). The authors of this study calculated how modifications on the features affect the outcome. The study reported AUC ROC = 0.77 (CI 95%, 0.72 to 0.82). Bernabeu-Wittel et al. in 2011 [2] proposed a method for detecting 1-year mortality for polypathological patients. That model computed the PROFUND score, based on some features to assign a mortality risk to the patient: Age, Hemoglobin, Barthel index, No caregiver or caregiver other than the spouse, hospital admissions >= 4 in last 12 months and positive for few diseases. The PROFUND score is mapped into mortality (in less than a year) probability. The reported validation result was AUC ROC = 0.7 (CI 95%, 0.67 to 0.74). Van Walraven et al. in 2015 [6] reported a 1-year mortality forecast model based on patient demographics, health burden, and severity of acute illness. The model uses a binomial logistic regression. The AUC ROC ranged from 0.89 (CI 95%, 0.87 to 0.91) to 0.92 (CI 95%, 0.91 to 0.92). Recently, Avati et al. in 2017 [13] presented a deep neural network for one-year mortality prediction by using 13654 features, corresponding to the different ICD9 codes in different time windows through the year. They reported and 0.93 of AUC ROC for all validation patients but only 0.87 for admitted patients.

Based on the promising results in the literature we have addressed the design of a high-performance predictive model of one-year mortality exclusively based on observations at hospital admission. The overall aim of our study was to provide quantitative methods to healthcare caregivers to decide the inclusion of patients in the palliative care program during the hospital admission. To this aim, we have designed and evaluated five predictive models from the state-of-the-art machine learning discipline. These models are meant to be in a complexity step between the first studies and the Avati's deep learning approximation, being the most adequate option for our dataset size.

## 2. Materials

The data of the study was extracted from the Electronic Health Records from Hospital La Fe. We gathered all the hospitalization episodes of adult patients (≥ 18 years old), excluding those related to mental health, gynecology and obstetrics, from January 2014 to December 2017 (a total number of 114393 cases) that have been discharged from the hospital. To guarantee independent observations, we selected a random single episode for each patient, reaching a total of 65279 episodes.

The dataset contains information about the previous and current admission (7 features), laboratory test results (7 features) and a list of 28 selected diseases for which the patient is positive or negative. Sex, age, Charlson index, and Barthel tests result are also available. This adds up a total of 36 features which can be obtained straightforwardly in the first hours of admission. Some of these features were used, with positive results, in previous studies.

Target variable was exitus after one year from the admission date. The number of patients that have died in less than a year (positive cases) was 8133 (~12.43%), the number of negative cases is 57166 (~87.57%). The whole variable description can be seen in Table I.

| Variable | Types | Missings | Distribution |
|---|---|---|---|
| Sex | CAT | 0 | Males: 51.86% |
| Age | INT | 0 | 61.327 ± 18.375 |
| Urgent Admission | BOOL | 0 | Pos: 56,97% |
| Admission Destination | CAT | 0 | - |
| Service | CAT | 0 | - |
| Admission Cause | CAT | 0 | - |
| Prev. Stays | INT | 0 | 6.119 ± 9.502 |
| Barthel Test | INT | 56214 | 67.268 ± 37.919 |
| Prev. Admissions | INT | 0 | 0.300 ± 0.789 |
| Prev. Emergency Room | INT | 0 | 0.935 ± 1.691 |
| Charlson Score | INT | 0 | 4.233 ± 3.238 |
| Albumin (g/dL) | REAL | 46857 | 2.955 ± 0.677 |
| Creatinine (mg/dL) | REAL | 16920 | 0.505 ± 1.063 |
| Hemoglobin (g/dL) | REAL | 14434 | 11.703 ± 2.228 |
| Leucocytes (Cel/mL) | REAL | 14434 | 9.457 ± 7.389 |
| PCR (mg/L) | REAL | 30285 | 63.083 ± 84.481 |
| Sodium (mEq/L) | REAL | 17183 | 139.672 ± 4.354 |

| Urea (mg/dL) | REAL | 18459 | 46.255 ± 34.628 |
|---|---|---|---|
| Acute Myocardial Infarction | BOOL | 0 | Pos: 3.09% |
| Congestive Heart Failure | BOOL | 0 | Pos: 6.14% |
| Peripheral Vascular Disease | BOOL | 0 | Pos: 4.88% |
| Cerebrovascular Disease | BOOL | 0 | Pos: 6.76% |
| Dementia | BOOL | 0 | Pos: 1.5% |
| Chronic Pulmonary Disease | BOOL | 0 | Pos: 10.03% |
| Rheumatic Disease | BOOL | 0 | Pos: 1.6% |
| Peptic Ulcer Disease | BOOL | 0 | Pos: 1.57% |
| Mild Liver Disease | BOOL | 0 | Pos: 5.77% |
| Diabetes Without Complications | BOOL | 0 | Pos: 13.19% |
| Diabetes With Complications | BOOL | 0 | Pos: 1.27% |
| Hemiplegia Paraplegia | BOOL | 0 | Pos: 1.28% |
| Renal Disease | BOOL | 0 | Pos: 7.46% |
| Malignancy | BOOL | 0 | Pos: 18.2% |
| Moderate Severe Liver Disease | BOOL | 0 | Pos: 1.49% |
| Metastasis | BOOL | 0 | Pos: 3.27% |
| AIDS | BOOL | 0 | Pos: 0.57% |
| Delirium | BOOL | 0 | Pos: 0.12% |

Table I: Features information.

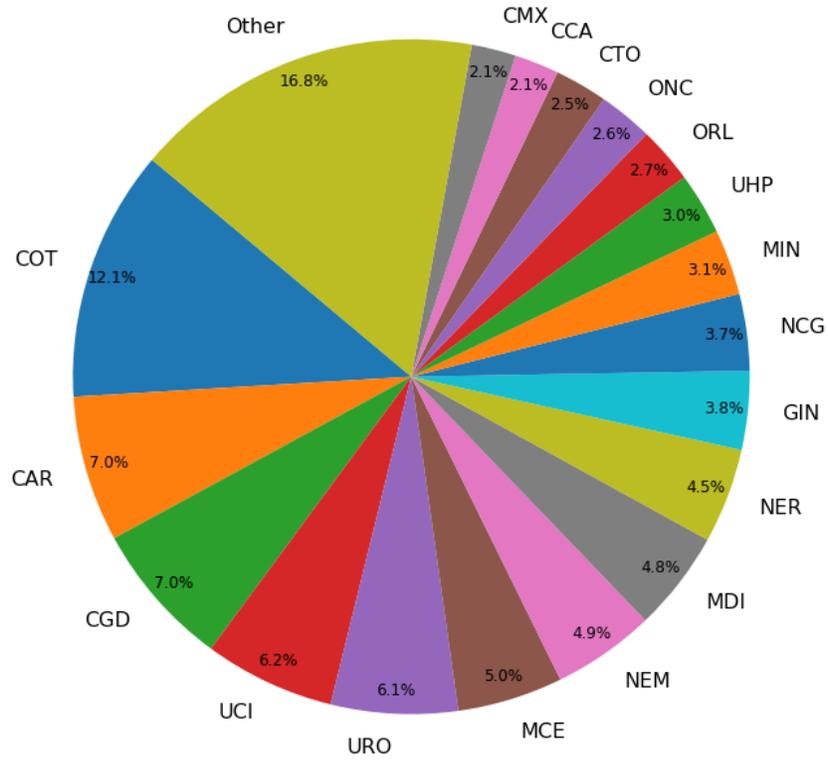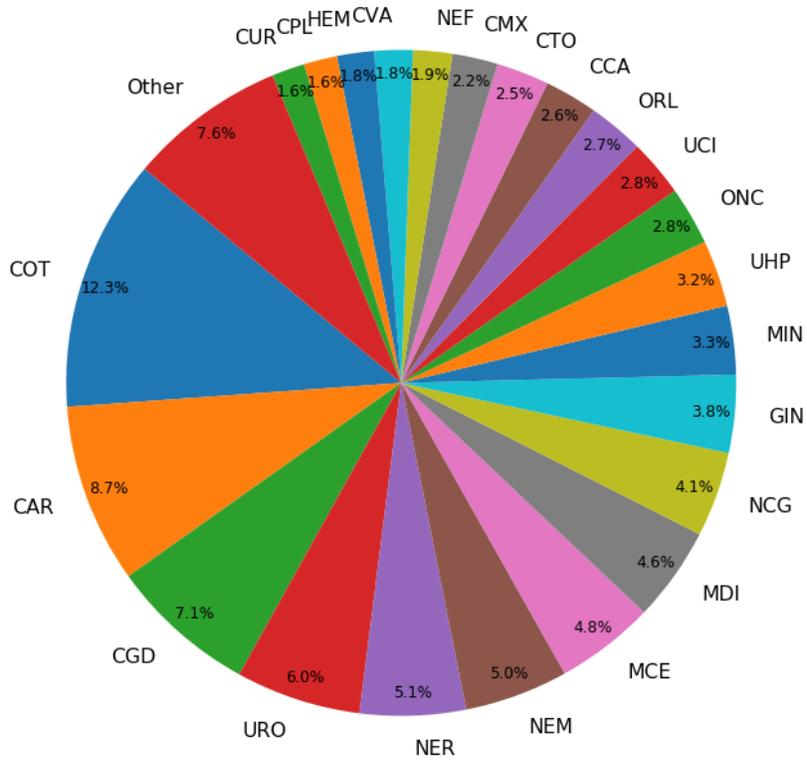

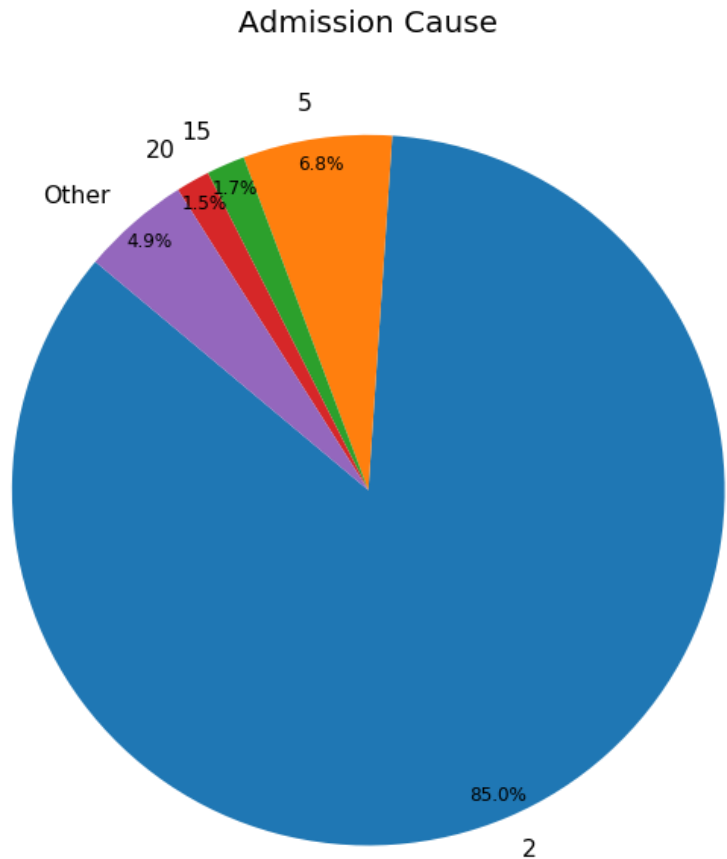

Figure 1: Distribution for the categorical features: Admission Destination, Service, Admission Cause.

## 3. Methods

### 3.1 Development of the models

Five machine learning techniques were employed for developing our predictive models: Gradient Boosting Classifier [7], Random Forest [17], K-Nearest Neighbors [5], Multilayer Perceptron (MLP) [16] and Support Vector Machine [15]. The implementation of the scikit-learn toolkit [3] was employed in all of them except in the MLP which uses Keras and TensorFlow [18]. Moreover, the optimization tool TPOT [4] was used in order to find a good model to fit the data.

### 3.2 Feature Importance

We studied the relevance of each feature for the final prediction by calculating the GINI importance provided by the Gradient Boosting Classifier. The GINI importance measures the average gain of purity by splits of a given variable. If the variable is discriminant for the problem, it tends to split mixed labeled nodes into pure single class nodes [20].

### 3.3 Validation of the State-of-the-Art models

As a first step, we have compared our model with the PROFUND and Buurman's model using the same evaluation method. For the Buurman's model, a clinical committee led by Vicente Ruiz-García at Hospital La Fe adapted the Buurman's proposal as one-year mortality index, using a linear regression with the 1-year mortality target variable. Besides, we evaluated the original PROFUND model proposed in [2]. The validation of the other models in state of the art was not possible due to the lack of part of their features in our data system.

### 3.4 Evaluation of the models

First, we have computed the ROC Curve [19] for each model and calculated the optimum probability threshold (minimum probability to assign the positive class to a sample) running our models using a random split of the data from separating train and test. We iterated over all the different values that could change the specificity and the sensitivity of the model and kept the threshold that minimizes the balanced error rate (BER) [19].

Once the threshold is established for each model, we internally validated them using a 100-repetition stratified hold-out (80% of the data in order to train the model and 20% for test it). The missing values have been imputed using the median of the train split. Five metrics have been stored for each experiment (accuracy, AUC ROC, specificity, sensitivity and balanced error rate) [19, 20]. For each metric, the mean and the 95% confidence interval have been computed.

## 4. Results

| Model | Threshold | Accuracy | AUC ROC | Specificity | Sensitivity | BER |
|---|---|---|---|---|---|---|
| Gradient Boosting Classifier | 0.1 | 0.813 [0.813, 0.814] | 0.911 [0.911, 0.912] | 0.807 [0.806, 0.808] | 0.858 [0.856, 0.860] | 0.168 [0.167, 0.169] |
| Random Forest | 1.725 | 0.824 [0.823, 0.824] | 0.902 [0.901, 0.903] | 0.823 [0.822, 0.824] | 0.829 [0.882, 0.831] | 0.174 [0.173, 0.175] |
| K-Nearest Neighbors | 0.08 | 0.742 [0.740, 0.743] | 0.868 [0.868, 0.869] | 0.726 [0.725, 0.728] | 0.848 [0.846, 0.851] | 0.213 [0.211, 0.214] |
| Support Vector Machine | 105 | 0.794 [0791, 0.797] | 0.858 [0.857, 0.859] | 0.798 [0.794, 0.801] | 0.767 [0.761, 0.772] | 0.218 [0.216, 0.219] |
| Multi-Layer Perceptron | 117 | 0.792 [0.784, 0.799] | 0.885 [0.884, 0.886] | 0.788 [0.778, 0.798] | 0.817 [0.807, 0.827] | 0.197 [0.196, 0.199] |
| Baseline: Buurman's modification | 1.010 | 0.794 [0.766, 0.821] | 0.681 [0.680, 0.683] | **0.838** **[0.804, 0.873]** | 0.483 [0.460, 0.505] | 0.340 [0.333, 0.346] |
| Baseline: PROFUND | 58 | 0.569 [0.534, 0.603] | 0.739 [0.738, 0.740] | 0.534 [0.493, 0.575] | 0.814 [0.800, 0.828] | 0.326 [0.313, 0.339] |

Table II: Complete results for the machine learning proposed models

| Variable | Importance (%) | Variable | Importance (%) |
|---|---|---|---|
| Service | 10.60 | Malignancy | 0.90 |
| Urea | 10.32 | Sex | 0.68 |
| Leucocytes | 8.65 | Congestive Heart Failure | 0.51 |
| PCR | 7.88 | Renal Disease | 0.42 |
| Age | 7.65 | Dementia | 0.40 |
| Creatinine | 6.99 | Chronic Pulmonary Disease | 0.35 |
| Albumin | 6.66 | Diabetes Without Complications | 0.34 |
| Prev. Stays | 5.83 | Acute Myocardial Infarction | 0.30 |
| Hemoglobin | 5.67 | Moderate Severe Liver Disease | 0.26 |
| Sodium | 5.07 | Cerebrovascular Disease | 0.25 |
| Charlson Score | 4.55 | Mild Liver Disease | 0.24 |
| Admission Destination | 3.65 | Peripheral Vascular Disease | 0.22 |
| Barthel Test | 3.20 | Rheumatic Disease | 0.21 |
| Prev. Emergency Room | 1.98 | Hemiplegia Paraplegia | 0.20 |
| Cause of admission | 1.71 | Peptic Ulcer Disease | 0.20 |
| Prev. Admissions | 1.70 | Diabetes with Complications | 0.17 |
| Urgent Admission | 1.10 | Delirium | 0.16 |

| Metastasis | 0.95 | AIDS | 0.05 |
|---|---|---|---|

Table III: Variable importance provided by GBC sorted by decreasing importance.

## 5. Discussion

The prediction of death before one-year could be a relevant criterion to admit the patients into palliative care programs [21]. Also, the prediction of the death at admission of the patient would help the hospital management to better manage its resources in a more accurate way.

We used the area under the ROC curve as the comparison metric because is the common metric to all other works in the SoA. We also chose the threshold for considering a sample into the positive class taking the value that minimizes the balanced error rate.

The Buurman's modified model and the PROFUND index have been validated, the models described in our work outperform them in terms of AUC ROC, sensitivity and specificity. Whereas our models presented a bigger number of features (36) than the mentioned articles (4 for Buurman's modified model and 9 for PROFUND index)

Comparing with the most recent work, Avati et al. 2017, that presented a neural network with 18 hidden layers of 512 neurons each was trained with 177011 patients. The models in our approach are trained with 52223 patients. The network used 13654 features as input, our model uses only 36. Finally, they achieved an AUC ROC of 0.93 for all their patients but it only achieved 0.87 when only admitted patients are considered. We achieved better results using a significative smaller amount features, this led to a more compact model that also is more interpretable since the best performing model is based on decision trees.

The best results in our models achieved the interval reported by van Walraven et al. 2015: 0.89-0.92 AUC ROC. Despite the number of final features is smaller in HOMR (10) two of their features are composed: "charlson comorbidity index score" (15 items) and "diagnostic risk score" (70 items), so at the end, HOMR requires more information about the patients than our models. The performance comparison with Avati et al. 2017 and HOMR have been made using their reported results which implies the use of different evaluations and datasets.

We obtained consistent results compared to other studies. In HOMR the features that are capable to add more points to the index are the admitting service (up to 28) and the 'age x comorbidity' (other 28 points). We agree with the most important variable (real service code) and the fifth one in importance order (age). Our second most important variable, nitrogen in urea, is included among the Buurman's model. Moreover, creatinine in blood is related to BUN and is a variable also associated with mortality is our results.

The clinical features included in our work have clinical relevance and appear in other clinical prediction rules. They appears in the records of our hospital databases in Spain and allow the creation of alerts for the clinicians to address patients, to palliative care programs not only for advanced oncology patients but for other chronic pathologies as dementia (A critical literature review exploring the challenges of delivering effective palliative care to older people with dementia, cardiac failure or chronic obstructive pulmonary disease (COPD) or end-stage renal disease (Comparing the Palliative Care Needs of Those With Cancer to Those With Common Non-Cancer Serious Illness) [23, 24, 25]

This study has caused a direct impact on Hospital La Fe since the model based on the Gradient Boosting Classifier has been implemented in the pre-production information systems and it is on a test stage. Once in the day, a program gathers all the admitted patients' data and extracts the features, this information is passed to the model who gives a posteriori probability and a label prediction, this information is stored on a separated table of the same database including the timestamp.

The main limitation of the study was the use of data from only one hospital, we can't ensure that the models learned with the study population are effective with patients of another country/region, or another type of hospital, Hospital La Fe is a tertiary Hospital a referral in the Valencia region, with different patients and severity.

In addition, the models only had an internal derivation, so we need to refine and validate this model to reproduce the findings with different settings (smaller hospital and with less severity illness) may be outside the same city or Valencian community. It is necessary to work on additional criteria for palliative care admission besides mortality, for example, introducing the available resources in the decision-making process. Also, an inclusion criterion for chronic patients is needed since their illness trajectories are different from other patients. [22]

## 6.. Conclusions

This work proposes machine-learning forecast of one-year exitus using data from hospital admission. Our forecast achieved an area under ROC curve of 0.9 and a BER of 0.17, being the Gradient Boosting Classifier the best model. The features used in the models correspond to basic demographic and administrative information, some laboratory results and a list of positives or negatives for certain diseases. The presented models could have an instant impact on every hospital, only the feature extraction module and the table for results need to be adapted to the particular information system of every hospital, the rest of the components are ready to set in production. Our results have reached the best results in the state-of-the-art, corresponding to the HOMR index which validation in few Canadian hospitals produces AUC ROC from 0.89 to 0.92.

## 7. Clinical Relevance Statement

The research showed that is possible to predict which patient will have a high risk of death before one year after hospital admission. These predictions were for a wide range of population, not only for the suspected patients with a short -life expectancy patients Palliative care must ensure the best quality of life at the end of life's patients.

## 8. Conflict of Interest

The authors declare that they have no conflicts of interest in the research.

## 9. Human Subjects Protections

This work did not imply any risk nor alteration in the habitual health care provided for the subjects. Only the authorized personnel could access the electronic records that were properly anonymized using dissociative and untraceable codes.

This investigation followed the International Guideline for Ethical Review of Epidemiological Studies [14] and the ethic committee from IIS LA FE that approved the study's protocol.

## 10. Summary Table

What was already known on the topic
- Different models for one-year mortality prediction, mostly based on scores
- Patients enrolled in early palliative care improve their quality of life compared with the ones receiving late palliative care or standard care.

What this study added to our knowledge
- One year forecast at hospital admission using Gradient Boosting Classifier and Random Forest reported the best results in state of the art, achieving performance feasible for clinical use.
- Only features gathered at the first hours of hospital admission may be enough as criteria for palliative care inclusion.
- Hospital service, laboratory analysis, and age are the essential features for one year forecast, confirming previous studies and allowing us to create compact predictive models

## 11. Acknowledgments

The authors acknowledge the collaboration for this article to the CrowdHealth project (COLLECTIVE WISDOM DRIVING PUBLIC HEALTH POLICIES (727560)), the MTS4up project (DPI2016-80054-R), the REDISEC program and the BDSLab investigators.